\documentclass[11pt]{article}

\usepackage[preprint]{acl}

\usepackage{times}
\usepackage{latexsym}
\usepackage[T1]{fontenc}
\usepackage[utf8]{inputenc}
\usepackage{microtype}
\usepackage{inconsolata}
\usepackage{graphicx}
\graphicspath{{figures/}}

\usepackage{booktabs}
\usepackage{amsmath}
\usepackage{amssymb}
\usepackage{multirow}
\usepackage{xcolor}
\usepackage{enumitem}
\usepackage{tikz}
\usetikzlibrary{positioning,arrows.meta,fit,backgrounds,calc}

\newcommand{\aic}{answer-in-context}
\newcommand{\Aic}{Answer-in-context}
\newcommand{\submod}{\texttt{chunk\_submod}}
\newcommand{\focused}{\texttt{chunk\_focused}}
\newcommand{\packed}{\texttt{chunk\_packed}}
\newcommand{\mmrpol}{\texttt{chunk\_mmr}}

\renewenvironment{abstract}{%
  {\centering\large\textbf{\abstractname}\par}\vspace{2pt}%
  \begin{list}{}{%
    \setlength{\rightmargin}{0.6cm}%
    \setlength{\leftmargin}{0.6cm}%
    \setlength{\topsep}{0pt}%
    \setlength{\partopsep}{0pt}%
    \setlength{\parsep}{0pt}%
    \setlength{\itemsep}{0pt}%
  }%
  \item[]\fontsize{10pt}{12pt}\selectfont\ignorespaces
}{%
  \unskip\end{list}%
}

\title{Recall Is Not Enough: A Reader-Context Diagnostic for\\
       Budget-Constrained Retrieval-Augmented Generation}

\author{Ananto Nayan Bala\\Ahsanullah University of Science and Technology\\ananto.cse.20210204028@aust.edu}

\begin{document}
\maketitle

\begin{abstract}
Retrieval-augmented generation under a fixed context budget forces a selection
problem: only a fraction of the retrieved evidence fits in front of the reader.
The field's standard metric, recall@$k$, is scored on the \emph{retrieved set},
but the reader consumes the \emph{packed context}---and once packing must
discard evidence, the two come apart.

We introduce \textbf{\aic{}}, a diagnostic that measures whether a gold answer
survives into the packed context, and argue it is the quantity budgeted RAG
should be optimizing. It carries substantial information beyond retrieval,
adding $\Delta R^2{=}0.17$--$0.27$ over recall across three multi-hop datasets;
even among questions where \emph{all} gold was retrieved, whether packing keeps
the answer separates exact match $4.6\times$. Two independent interventions
confirm the mediation: a packing change that raises document coverage without
raising \aic{} leaves accuracy flat, and prompt compression that destroys the
answer span lowers both together. A graded variant extends the diagnostic to
free-form answers, where no verbatim span exists.

We then show the diagnostic is actionable. Casting reader-context construction
as budgeted submodular maximization gives a packer that beats both deployed
top-$k$ truncation and LLMLingua-2 compression---across three reader families,
four scales, and four budgets, at equal-or-lower token cost. Against a
hand-tuned query-focused heuristic, which we show approximates the same
objective, it reaches parity, winning outright only where evidence density is
the binding constraint. Throughout, one variable predicts what helps and what
cannot.
\end{abstract}

\section{Introduction}
\label{sec:intro}

A retrieval-augmented reader has a finite context window, and in practice an
even smaller \emph{evidence budget}: the share of that window allocated to
retrieved passages. Once retrieval returns more relevant text than fits, the
system must decide what to keep. This selection step is usually treated as an
afterthought---concatenate the top-$k$, truncate to fit
\citep{lewis2020retrieval,ram2023incontext}---yet under a tight budget it is the
step that decides whether the reader ever sees the answer.

\begin{figure}[t]
\centering
\begin{tikzpicture}[
  box/.style={draw,rounded corners=1.5pt,minimum height=5mm,inner sep=3pt,align=center,font=\scriptsize},
  proc/.style={draw,fill=blue!8,rounded corners=2pt,minimum height=6mm,inner sep=3pt,align=center,font=\scriptsize\bfseries},
  setbox/.style={draw,dashed,rounded corners=2pt,inner sep=1.4pt},
  gold/.style={draw,fill=green!20,minimum width=3.4mm,minimum height=3mm,inner sep=0pt},
  dist/.style={draw,fill=black!10,minimum width=3.4mm,minimum height=3mm,inner sep=0pt},
  lbl/.style={font=\scriptsize\itshape,text=black!60,align=center},
  ar/.style={-{Stealth[length=4pt]}},
]
\node[box] (q) {query};
\node[proc,right=4mm of q] (ret) {Retriever};
\node[setbox,right=6mm of ret,minimum width=20mm,minimum height=5.5mm] (retset) {};
\node[gold] at ([xshift=-7.2mm]retset.center) (g1) {};
\node[gold] at ([xshift=-3.0mm]retset.center) (g2) {};
\node[dist] at ([xshift= 1.2mm]retset.center) (d1) {};
\node[dist] at ([xshift= 5.4mm]retset.center) (d2) {};
\node[lbl,above=0.4mm of retset.north] {retrieved set\ \ (recall@$k$ here)};
\draw[ar] (q) -- (ret);
\draw[ar] (ret) -- (retset);
\node[proc,below=7mm of retset] (pack) {Packer\\($\le B$ tokens)};
\draw[ar] (retset.south) -- (pack.north);
\node[setbox,left=6mm of pack,minimum width=11mm,minimum height=5.5mm] (ctx) {};
\node[gold] at ([xshift=-2.4mm]ctx.center) (cg) {};
\node[dist] at ([xshift= 2.4mm]ctx.center) (cd) {};
\node[lbl,below=0.4mm of ctx.south] {context (\aic{} here)};
\node[font=\scriptsize,text=red!70,above=1.4mm of ctx,align=center]
      (drop) {gold\,\#2 dropped};
\node[proc,below=7mm of ctx] (rd) {Reader};
\node[box,left=4mm of rd] (ans) {answer};
\draw[ar] (pack) -- (ctx);
\draw[ar] (ctx.south) -- (rd.north);
\draw[ar] (rd) -- (ans);
\end{tikzpicture}
\caption{Recall is scored on the \emph{retrieved set}; the reader consumes the
\emph{packed context}. Under a budget the packer can drop a retrieved gold
document (here ``gold \#2''), so high recall need not mean the answer survives.
\Aic{} measures exactly what reaches the reader.}
\label{fig:pipeline}
\end{figure}
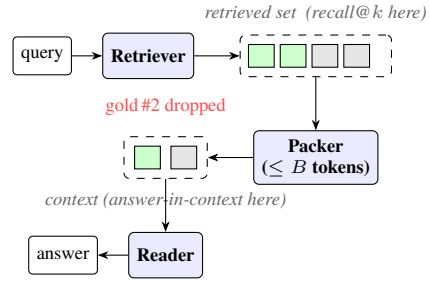

The community's default retrieval metric, recall@$k$, is computed on the
\emph{retrieved document set}. But the reader never consumes the retrieved set;
it consumes the \emph{packed context}. When packing discards evidence to fit a
budget, recall and what-the-reader-sees diverge. The divergence is acute for
\textbf{multi-hop} questions \citep{yang2018hotpotqa,trivedi2022musique}, where
the answer depends on combining evidence from several documents: retrieving all
of them is necessary but not sufficient, because the packer may keep a redundant
pair and drop the bridge. Figure~\ref{fig:pipeline} makes the gap concrete.

This paper is primarily about \emph{measurement}. We ask what property of the
reader context actually predicts answer quality under a budget, and answer with
\textbf{\aic{}}: does a gold answer survive into the packed context? It predicts
answer F1 far better than retrieval recall on every dataset we test, and---more
importantly---carries information recall does not have, on three multi-hop
datasets, with two independent interventions confirming it mediates the effect
of packing on accuracy (\S\ref{sec:diagnostic}). This reframes the budgeted-RAG
objective from ``retrieve the gold documents'' to ``pack so the answer
survives.''

A diagnostic is only useful if it is \emph{actionable}, so the second half of
the paper asks whether optimizing it directly helps. We formulate reader-context
construction as \textbf{budgeted monotone submodular maximization}
(\S\ref{sec:method}) and find that it does: the resulting packer beats both the
top-$k$-and-truncate packing deployed in practice and a strong prompt-compression
baseline, across three reader families, four scales, and four budgets, at
equal-or-lower token cost (\S\ref{sec:results}). A per-question decomposition
ties the gain to the diagnostic---the packer helps precisely by assembling
complementary multi-hop evidence into the context.

Against a \emph{hand-tuned query-focused heuristic}, however, the packer only
reaches parity. We take this seriously rather than bury it (\S\ref{sec:scope}).
That heuristic turns out to approximate the same objective by hand, which
explains both the parity and the one regime where the principled version pulls
ahead: a small reader under a binding budget on complementary evidence, where
density is what limits accuracy. Everywhere else---larger readers, single-pass
tasks, retrieval-bottlenecked ones---the two are statistically indistinguishable,
and the diagnostic says why in each case.

\paragraph{Contributions.}
\begin{enumerate}[noitemsep,topsep=2pt,leftmargin=1.4em]
  \item \textbf{A diagnostic, and its validation.} \Aic{}, a
    reader-context-level metric that predicts budgeted-RAG quality better than
    recall, with \emph{incremental validity} over recall on three multi-hop
    datasets ($\Delta R^2{=}{+}0.17$ to ${+}0.27$; a $4.6\times$ EM separation
    that survives even when all gold is retrieved), \emph{two independent
    interventional} confirmations, and a graded variant extending it to
    free-form answers.
  \item \textbf{Evidence that it is actionable.} A budgeted submodular packer
    that beats deployed top-$k$ packing and LLMLingua-2 compression across three
    reader families, 3B--32B readers, and four budgets, with a mechanistic
    per-question explanation tying the gain to \aic{}.
  \item \textbf{An honest account of the harder comparison.} We show a
    hand-tuned heuristic approximates our objective, report that it and the
    packer are indistinguishable in most settings, and locate the regime where
    principled packing wins outright---tied throughout to the diagnostic.
\end{enumerate}

The result is a validated measurement for budgeted RAG plus evidence that acting
on it pays, stated at the strength the data supports rather than inflated into a
universal claim that would not survive replication.

\section{Related Work}
\label{sec:related}

\paragraph{Retrieval-augmented generation.}
RAG couples a (typically dense; \citealp{karpukhin2020dense}) retriever with a
reader LM
\citep{lewis2020retrieval,guu2020realm,izacard2021leveraging,izacard2023atlas}
and now spans retrieval from trillions of tokens \citep{borgeaud2022improving},
in-context retrieval \citep{ram2023incontext}, black-box augmentation
\citep{shi2024replug}, joint instruction tuning \citep{lin2024radit}, and
self-reflective variants \citep{asai2024selfrag}; see \citet{gao2023retrieval}
for a survey. Most of this
work reports retrieval recall and end-task accuracy \emph{separately} and treats
context construction as fixed top-$k$ concatenation. Our diagnostic targets the
quantity in between---what the packed context actually contains---which becomes
the binding variable once a budget forces selection.

\paragraph{Multi-hop question answering.}
HotpotQA \citep{yang2018hotpotqa}, MuSiQue \citep{trivedi2022musique},
2WikiMultiHopQA \citep{ho2020constructing}, and WikiHop
\citep{welbl2018constructing} require composing evidence across documents.
A large line of work attacks the retrieval side of this difficulty with
multi-hop dense retrieval \citep{xiong2021answering}, interleaved
retrieval-and-reasoning \citep{trivedi2023interleaving,press2023measuring} built
on chain-of-thought prompting \citep{wei2022chain}, iterative
retrieval-generation \citep{shao2023enhancing,jiang2023active}, and
program-style composition \citep{khattab2022demonstrate}. We use these datasets
not to improve retrieval but to \emph{vary} whether the complementary evidence
is present and surfaced, which is what determines whether a packer can help.

\paragraph{Context selection and compression.}
Reducing reader context via reranking, selection, or compression is well
studied. The canonical redundancy-aware reranker is Maximal Marginal Relevance
(MMR) \citep{carbonell1998use}, our direct baseline. Recent methods compress or
filter retrieved context---RECOMP \citep{xu2024recomp}, LLMLingua
\citep{jiang2023llmlingua}, Selective Context \citep{li2023compressing},
context filtering \citep{wang2023learning}, and robustness to irrelevant
passages \citep{yoran2024making}. ``Lost in the middle'' effects
\citep{liu2024lost} and long-context studies
\citep{bai2024longbench,xu2024retrieval} show that simply enlarging the window
is not a substitute for choosing what goes in it. Our packer differs in that its
objective is tied to an explicit, measurable answer-density quantity (the
diagnostic), and our central message is a scope map for \emph{when} principled
selection helps at all. We also run LLMLingua-2 \citep{pan2024llmlingua2} as a
live baseline (\S\ref{sec:results}) rather than only citing this line, and find
selection beats compression decisively under a tight budget.

\paragraph{Submodular optimization for selection.}
Coverage-and-diversity objectives with the cost-scaled greedy algorithm and
its constant-factor guarantee \citep{nemhauser1978analysis} were introduced for
extractive summarization by \citet{lin2011class,lin2010multidocument}; see
\citet{krause2014submodular,bilmes2022submodularity} for broader treatments.
We apply that machinery to \emph{reader-context evidence packing} for RAG and
tie the objective to the answer-in-context quantity our diagnostic measures.

\paragraph{Retrievers and readers.}
We use a bi-encoder retriever \citep{reimers2019sentence,xiao2024cpack} of the
kind evaluated on MTEB \citep{muennighoff2023mteb} and BEIR
\citep{thakur2021beir}, with classic sparse \citep{robertson2009probabilistic},
late-interaction \citep{khattab2020colbert}, and cross-encoder
\citep{nogueira2019passage} retrieval as the surrounding context. Readers are
instruction-tuned LLMs \citep{qwen2025qwen25,touvron2023llama,brown2020language};
the larger rungs of our reader ladder use 4-bit NF4 quantization
\citep{dettmers2023qlora,dettmers2022llmint8} to fit commodity GPUs, which is
why we include a precision control.

\paragraph{RAG evaluation.}
EM/F1 \citep{rajpurkar2016squad} measure answer quality, while RAG-specific
frameworks score faithfulness and context relevance
\citep{es2024ragas,saadfalcon2024ares,chen2024benchmarking} over knowledge-
intensive suites \citep{petroni2021kilt,mallen2023trust}. These score the
\emph{retrieved} context or the \emph{final} answer; \aic{} instead measures the
packed context the reader sees, and we show it has incremental validity over
recall for predicting end-task quality.

\section{The Answer-in-Context Diagnostic}
\label{sec:diagnostic}

\subsection{Definition}
\label{ssec:def}

Given a question with gold answer set $A$ and a \emph{materialized reader
context} $C$ (the concatenation of packed snippets actually shown to the
reader), we define:
\begin{itemize}[noitemsep,topsep=2pt,leftmargin=1.2em]
  \item \textbf{\aic{}} ${=}1$ if some normalized $a\in A$ occurs as a contiguous
    token subsequence of normalized $C$, else $0$;
  \item \textbf{gold-doc reader coverage}: fraction of gold documents
    contributing $\ge 1$ snippet to $C$; \textbf{all-gold-in-reader}: whether
    \emph{all} of them do;
  \item \textbf{gold-token density}: fraction of $C$'s tokens drawn from gold
    documents.
\end{itemize}
These are computed on the \emph{packed} run, not the retrieved set---the key
difference from recall@$k$, which is scored on retrieved document ids
\emph{before} packing. \Aic{} is a necessary condition for an extractive-style
reader to be correct, and we hypothesize it is the mediator explaining why higher
recall need not raise answer quality under a budget.

\subsection{\Aic{} predicts quality; recall does not}
\label{ssec:correlations}

\begin{table}[t]
\centering\small
\begin{tabular}{lcc}
\toprule
Feature & corr.\ w/ F1 & corr.\ w/ EM \\
\midrule
\textbf{\aic{}}         & \textbf{+0.50} & \textbf{+0.46} \\
gold-doc reader cov.    & +0.33          & +0.29 \\
retrieval all-gold@5    & +0.32          & +0.27 \\
retrieval recall@5      & +0.31          & +0.27 \\
gold-token density      & +0.26          & +0.23 \\
\bottomrule
\end{tabular}
\caption{Feature--quality correlations on HotpotQA (seed 42, 500 questions,
$n{=}2{,}500$ policy$\times$question rows, budget 160). \Aic{} is the strongest
single predictor---above both retrieval metrics.}
\label{tab:correlations}
\end{table}

Table~\ref{tab:correlations} pools all policy$\times$question rows on HotpotQA and
correlates each diagnostic with answer quality. \Aic{} is the strongest single
predictor, above both retrieval metrics and reader-level coverage. Conditioning
directly: mean F1 is $0.596$ when a gold answer is in the reader context versus
$0.123$ when it is not (a $+0.47$ gap). This resolves the ``lower recall, better
answers'' paradox: under a budget, what matters is whether the answer
\emph{survives into context}, not how many gold documents were retrieved.

\subsection{Incremental validity: not recall in disguise}
\label{ssec:incremental}

\begin{figure}[t]
\centering
\includegraphics[width=0.86\columnwidth]{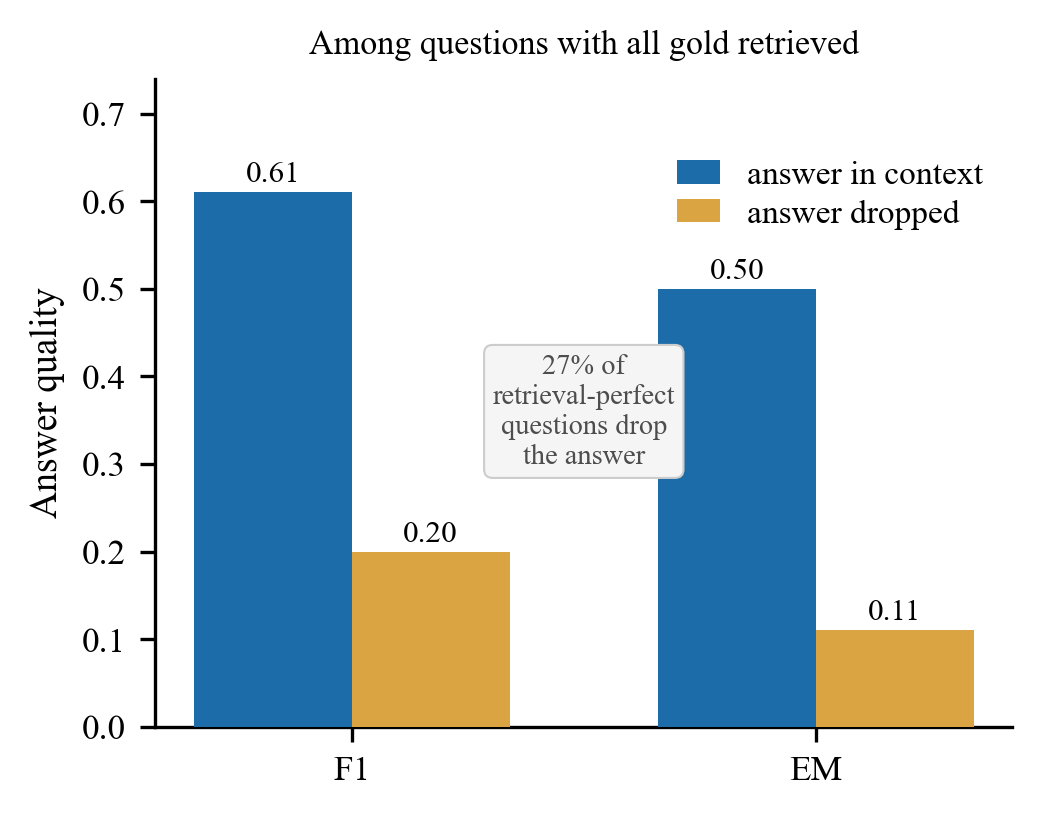}
\caption{Among HotpotQA questions where \emph{all} gold paragraphs were
retrieved (recall@5${=}1$), whether packing keeps the answer in context is still
decisive: F1 $0.61$ vs.\ $0.20$, EM $0.50$ vs.\ $0.11$. $27\%$ of these
retrieval-perfect questions drop the answer during packing. Clustered bootstrap
on question id, three seeds.}
\label{fig:aic}
\end{figure}

A natural objection is that \aic{} is near-tautological with correctness, or a
proxy for recall. Two analyses refute this, pooling per-question rows across
three seeds $\{42,13,7\}$ (10,500 rows per dataset) with inference
\emph{cluster-robust on question id}.

\textbf{(a) Incremental validity over recall.} On HotpotQA, a model of F1 on
recall@5 alone explains $R^2{=}0.085$; adding \aic{} raises this to
$R^2{=}0.257$, an increment of $\Delta R^2{=}{+}0.17$ $[+0.15,+0.20]$. The
standardized coefficient on \aic{} ($\beta{=}{+}0.46$) is ${\approx}4\times$ that
on recall ($\beta{=}{+}0.10$), and the partial correlation of \aic{} with F1
controlling for recall is $+0.43$. \Aic{} and recall@5 themselves correlate only
$+0.41$---far from the ${\approx}1$ that ``it is just recall'' would require.
The same analysis on the other two multi-hop datasets yields a \emph{larger}
increment: MuSiQue $\Delta R^2{=}{+}0.23$ $[+0.17,+0.29]$ and 2Wiki
$\Delta R^2{=}{+}0.27$ $[+0.24,+0.31]$, with \aic{}-to-recall coefficient ratios
of $6\times$ and $12\times$ and \aic{}--recall correlations of only $+0.37$ and
$+0.24$. The diagnostic's advantage over recall is thus not a HotpotQA
artifact---it is largest exactly where retrieval is least informative.

\textbf{(b) It captures the packing step, orthogonal to retrieval.} Restrict to
questions where retrieval already succeeded---all gold in the top-5
($n{=}7{,}739$). Even here, \textbf{27\% still drop the answer during packing}
(Figure~\ref{fig:aic}). Within this retrieval-perfect subset, whether packing
keeps the answer is decisive: F1 $0.61$ vs.\ $0.20$ and EM $0.50$ vs.\ $0.11$
(a $3.0\times$/$4.6\times$ gap, tight clustered-bootstrap CIs). This is the
cleanest evidence that \aic{} measures the \emph{packing} step rather than
restating retrieval or correctness. (A minority of the 27\% are paraphrased
answers, not packing failures, modestly overstating packing's share; the
predictive-validity conclusion is unaffected.)

\subsection{Generalization and an interventional test}
\label{ssec:generalization}

\begin{table}[t]
\centering\small
\setlength{\tabcolsep}{4pt}
\begin{tabular}{@{}p{2.45cm}cp{1.85cm}@{}}
\toprule
Dataset & \shortstack{$r$(\aic{},\\ F1)} & Note \\
\midrule
HotpotQA          & \textbf{+0.50} & packer wins \\
MuSiQue           & \textbf{+0.54} & packer null \\
2WikiMultiHopQA   & \textbf{+0.55} & packer null; interventional \\
RAGBench CovidQA  & +0.39          & single-pass \\
RAGBench ExpertQA & degenerate     & long free-form answers \\
\bottomrule
\end{tabular}
\caption{Verbatim \aic{}--F1 correlation across five datasets. Strongest on the
two datasets where the packer shows no win---not an artifact of the method.
Degenerate on ExpertQA (answers never appear verbatim); the graded variant
recovers it ($r{=}0.32$, \S\ref{ssec:generalization}).}
\label{tab:generalization}
\end{table}

Table~\ref{tab:generalization} shows the correlation is not specific to HotpotQA
or to our packer; it is in fact strongest on MuSiQue and 2Wiki, where the packer
shows no win. This is the key evidence that the diagnostic is a
dataset-independent mediator, not a side effect of the method.

\paragraph{An interventional test on 2Wiki.}
\S\ref{ssec:incremental} is observational; 2WikiMultiHopQA lets us test the
diagnostic \emph{interventionally}. Running the exact HotpotQA factorial on
2Wiki (3B reader, budget 160, three seeds), the packer assembles strictly more
gold than the focused heuristic---gold-doc coverage $+0.054$ on all three
seeds---yet moves \aic{} by $-0.007$ and F1 by $-0.008$ ($p{=}0.44$, a clean
null): on 2Wiki's compositional questions the answer usually sits in the
document the heuristic already ranks first, so the extra gold is bridging
evidence that scaffolds reasoning without carrying the answer string. Move
coverage but not \aic{}, and quality does not move (details in
App.~\ref{app:2wiki}). \S\ref{sec:results} supplies the converse intervention:
compression that \emph{lowers} \aic{} lowers accuracy with it.

\paragraph{A graded variant for free-form answers.}
The verbatim-span check is binary, and degenerate where an answer never appears
as a contiguous run---long free-form responses (RAGBench ExpertQA), where it is
$0$ for every question and its F1 correlation is undefined. We therefore also
compute a \textbf{graded \aic{}}: the fraction of gold-answer tokens within a
local window (width $2{\times}$ the answer length) of the context, maximized over
answers. It equals the binary check wherever a span exists but stays informative
when the answer is long. On span datasets it tracks the binary version (HotpotQA
$r{=}0.50$ vs.\ $0.49$; MuSiQue $0.56$ vs.\ $0.54$); on free-form ones it
\emph{recovers} the lost signal (CovidQA $0.56$ vs.\ $0.39$; ExpertQA $0.32$,
where the binary version is undefined). The diagnostic thus extends past
extractive QA at no extra model cost.

\section{Method: Budgeted Submodular Evidence Packing}
\label{sec:method}

\subsection{Objective}
\label{ssec:objective}

Given retrieved evidence for a query and a hard reader-token budget $B$, we build
a candidate set of source-grounded snippets and select a subset $S$ maximizing
\begin{equation}
\label{eq:objective}
\begin{split}
F(S) ={}& w_{\mathrm{rel}}\,\mathrm{Rel}(S) + w_{\mathrm{qry}}\,\mathrm{QueryCov}(S)\\
        & + w_{\mathrm{cov}}\,\mathrm{Repr}(S) + w_{\mathrm{div}}\,\mathrm{Div}(S)
\end{split}
\end{equation}
subject to $\mathrm{cost}(S)\le B$ and a snippet cap. Each term is monotone and
submodular, normalized to $[0,1]$:
\textbf{Rel} (modular) is the same per-snippet lexical relevance the focused
heuristic uses---so heuristic and packer see identical candidates and singleton
scores, isolating the \emph{selection rule}; \textbf{QueryCov} is a set-cover
over distinct query content terms; \textbf{Repr} is a saturated
facility-location term, $\sum_i \min\!\big(\sum_{j\in S}\mathrm{sim}(i,j),\,
\alpha\deg_i\big)$, that rewards covering candidate mass but saturates so it
cannot be gamed by near-duplicates; \textbf{Div} is a concave-over-documents
term, $\sum_d\sqrt{\text{relevance mass of }S\text{ in }d}$, spreading selection
across sources. We lead with relevance ($w_{\mathrm{rel}}{=}1.0$,
$w_{\mathrm{qry}}{=}0.5$, $w_{\mathrm{cov}}{=}0.4$, $w_{\mathrm{div}}{=}0.3$,
$\alpha{=}0.3$); the other three terms act as coverage/redundancy regularizers
that push complementary, answer-bearing evidence into the budget.

\subsection{Algorithm}
\label{ssec:algorithm}

We maximize $F$ with \textbf{cost-scaled (per-token) greedy}---at each step add
the feasible snippet with the largest marginal-gain-per-token ratio---followed
by the \textbf{Lin--Bilmes singleton fallback}: if the single best feasible
snippet outscores the greedy set, return it instead. This is the standard greedy
template for budgeted (knapsack-constrained) monotone submodular maximization
\citep{lin2010multidocument,lin2011class}; stronger constant-factor guarantees
for the knapsack case require additional partial enumeration
\citep{sviridenko2004note,nemhauser1978analysis}, which we do not perform---we
use the algorithm for its empirical behaviour under a token budget, not for a
guarantee. The contribution is not the
optimizer (textbook) but (a)~applying it to reader-context packing,
(b)~the four-term objective tied to answer density, and (c)~the controlled
evaluation isolating the selection rule from the candidate features.

\subsection{Baselines and the factorial}
\label{ssec:baselines}

Every packer consumes the \emph{same} candidates, so comparisons isolate the
objective. \textbf{Naive packed}: greedily concatenate by relevance until the
budget is hit---the common top-$k$-and-truncate strategy assumed by standard RAG
pipelines \citep{lewis2020retrieval,ram2023incontext}, and therefore a
practically meaningful baseline rather than a strawman.
\textbf{MMR} \citep{carbonell1998use}:
$\arg\max_i[\lambda\,\mathrm{rel}(i)-(1{-}\lambda)\max_{j\in S}\mathrm{sim}(i,j)]$,
$\lambda{=}0.7$---the natural ``isn't this just redundancy reduction?'' control.
\textbf{LLMLingua-2} \citep{pan2024llmlingua2} compresses the same candidates to
the same budget---the ``why not just compress?'' control.

\paragraph{The focused heuristic approximates our objective.}
The strongest baseline needs a word of its own, because it is not an independent
competitor. The \textbf{focused heuristic} (this project's prior best packer)
greedily prefers snippets contributing \emph{new query-term coverage}
\emph{across distinct documents}. Those two preferences are hand-designed
analogues of two terms in Eq.~\ref{eq:objective}: query-term coverage is a
crude $\mathrm{QueryCov}$, and spreading across documents is a crude
$\mathrm{Div}$. What it lacks is the rest---no representativeness term, no
saturation, no cost normalization (it never divides gain by length), and no
approximation guarantee, since it checks the budget only after the fact. It is
best read, then, as a hand-tuned greedy approximation of the same objective.
This matters for interpreting \S\ref{sec:results}: where the two tie, the
natural reading is not that the objective fails but that the heuristic already
captures most of it, and the principled formulation earns its keep by deriving
those preferences rather than hand-setting them---and by supplying the missing
budget-awareness. Because the same packers also apply to \textbf{ACE}
graph-structured evidence (a source-linked claim/entity graph from earlier
project stages), we run the full $\{$chunk, ACE$\}$ factorial and report it in
App.~\ref{app:seed42}.

\section{Results: Does Optimizing the Diagnostic Help?}
\label{sec:results}

\paragraph{Setup.}
All runs share a pipeline: \texttt{bge-small-en-v1.5} embeddings truncated to
320 dimensions, \texttt{Qwen2.5-3B-Instruct} reader, on dual T4 GPUs.
HotpotQA uses 500 questions; the headline is replicated across seeds
$\{42,13,7\}$. The primary budget is 160 reader tokens. Significance is paired
bootstrap (10,000 resamples, 95\% CI); multi-seed tests pool (seed, question)
instances.

\begin{table}[t]
\centering\small
\begin{tabular}{lccc}
\toprule
Policy & F1 & EM & Tokens \\
\midrule
\texttt{llmlingua} (compr.) & 0.326 & 0.233 & 129.6 \\
\packed{} (naive)        & 0.400 & 0.306 & 151.1 \\
\focused{}               & 0.429 & 0.331 & 152.1 \\
\mmrpol{}                & 0.410 & 0.313 & 151.7 \\
\textbf{\submod{}}       & \textbf{0.451} & \textbf{0.359} & \textbf{145.5} \\
\midrule
oracle (mixed)           & 0.601 & 0.487 & 141.5 \\
\bottomrule
\end{tabular}
\caption{Three-seed means, HotpotQA-500, budget 160, 3B reader. \submod{} is the
best fixed policy on \emph{every} seed, at \emph{fewer} tokens; LLMLingua-2
compression of the same candidates is far behind.}
\label{tab:hotpotqa}
\end{table}

\paragraph{Optimizing \aic{} beats deployed practice.}
In Table~\ref{tab:hotpotqa}, \submod{} is the best fixed policy on every seed, at
\emph{fewer} tokens (${\approx}145$ vs.\ ${\approx}152$). Against \packed{}---the
top-$k$-and-truncate packing standard pipelines assume---it wins by $+0.051$ F1
$[+0.030,+0.072]$ (three-seed bootstrap, $n{=}1{,}500$), at \emph{lower} cost
rather than more context. Against the hand-tuned \focused{} heuristic it is ahead
here by $+0.022$ $[+0.002,+0.041]$, an advantage that---as
\S\ref{sec:scope} shows---does not hold generally, and which we interpret in
light of that heuristic approximating the same objective
(\S\ref{ssec:baselines}). Plain MMR is \emph{significantly worse} than
\focused{} ($-0.020$ F1), so generic redundancy reduction hurts: only the full
coverage+representativeness+diversity objective helps, which settles the ``isn't
this just MMR?'' question. Running the same packers over a graph-structured
representation (ACE, from earlier project stages) does not help either---the
graph already de-duplicates, leaving little redundancy to exploit
(App.~\ref{app:seed42})---so what matters is the packing objective, not the
representation.

\paragraph{Against a real compression baseline.}
Selection is not the only way to fit a budget: the compression line
\citep{xu2024recomp,jiang2023llmlingua} shortens the context instead. We run
LLMLingua-2 \citep{pan2024llmlingua2} on the \emph{same} candidate pool at the
\emph{same} budget, so the only difference from \submod{} is the mechanism.
Compression loses badly---\submod{} beats it by $+0.125$ F1
$[+0.101,+0.149]$, and even \emph{naive} packing beats it by $+0.075$
$[+0.051,+0.098]$ (both $p{<}0.001$).\footnote{Not a budget artifact: LLMLingua
lands at ${\approx}130$ tokens, and naive packing at $B{=}128$ still scores
$0.368$ vs.\ its $0.326$ (Table~\ref{tab:app_budget}), so it loses even
token-matched.} The diagnostic says why: token-level dropping destroys the
answer span, cutting \aic{} by $0.243$ $[0.215,0.271]$ relative to \submod{}.
This is a third, independent confirmation of the mediation---an intervention
that lowers \aic{} lowers accuracy---and it is the sharpest practical statement
in the paper: under a tight budget, \emph{which} evidence you keep matters more
than how tightly you compress it.

\paragraph{Mechanism: complementary multi-hop assembly.}
A per-question decomposition (seed 42) attributes \textbf{81\%} of the
submod--focused gain to \textbf{37 questions} where the packer \emph{newly placed
a gold answer into the reader context} (${\approx}{+}0.39$ F1 each). The route is
better complementary coverage---all gold documents reach the context on 289
questions under submod vs.\ 256 under focused---not higher raw token density.
The packer wins by moving exactly the quantity the diagnostic measures. These
results use a 3B reader; \S\ref{ssec:reader} shows the advantage \emph{over the
focused heuristic} is specific to this scale, while the win over naive packing
and the mechanism persist.

\paragraph{Cross-family robustness.}
To check the win is not a Qwen artifact, we re-ran the HotpotQA budget-160
factorial (three seeds) on two other ${\approx}3$B families,
\texttt{Falcon3-3B-Instruct} \citep{falcon3} and
\texttt{Phi-3.5-mini-instruct} \citep{abdin2024phi3}
(Table~\ref{tab:crossfamily}); it separates cleanly. The win over
naive packing replicates and is significant on both (Falcon $+0.029$,
$p{=}0.01$; Phi $+0.048$, $p{<}0.001$); the narrower edge over the focused
heuristic does not (Falcon $-0.004$; Phi $+0.002$, both null), so that edge is
Qwen-3B-specific. This is not a failure but another boundary of the same
conditional claim the scope map (\S\ref{sec:scope}) is built around---the part
that generalizes is the win over naive packing.

\begin{table}[t]
\centering\small
\setlength{\tabcolsep}{5pt}
\begin{tabular}{lcc}
\toprule
Reader family (3B) & submod$-$naive & submod$-$focused \\
\midrule
Qwen2.5-3B (\S\ref{sec:results}) & $+0.051^{*}$ & $+0.022^{*}$ \\
Falcon3-3B                       & $+0.029^{*}$ & $-0.004$ \\
Phi-3.5-mini                     & $+0.048^{*}$ & $+0.002$ \\
\bottomrule
\end{tabular}
\caption{Cross-family replication on HotpotQA (budget 160, pooled three-seed
paired bootstrap, $n{=}1{,}500$). $^{*}${:}~$p{<}0.05$. The win over naive
packing is significant on all three families; the edge over the focused
heuristic is Qwen-specific.}
\label{tab:crossfamily}
\end{table}

A per-question oracle reaches F1${\approx}0.60$ vs.\ the best fixed policy's
${\approx}0.45$; since \submod{} is already (tied-)best on $79\%$ of questions
and the deciding variable on the rest is unobservable at inference time, we
report the oracle as headroom, not a deployed method.

\section{When Does Principled Packing Help? A Scope Map}
\label{sec:scope}

The wins over naive packing and over compression hold throughout this section.
What is \emph{located}, not universal, is the tighter comparison against the
focused heuristic, and two things bound it: whether there is complementary
evidence to assemble at all, and how capable the reader is.

\paragraph{Stating the comparison honestly.}
Before mapping the boundary we should be plain about its overall shape. Counting
every submod--focused contrast in this paper---one per dataset, budget, reader
scale, and reader family---gives \textbf{14 comparisons, of which 12 are
statistically indistinguishable}; the exceptions are HotpotQA at $B{=}160$ with a
3B reader ($+0.022$, in favour of the packer) and the 14B rung ($-0.029$, against
it). Read as a family, the honest summary is that the principled packer and the
hand-tuned heuristic \emph{perform comparably in most settings}, which is what
\S\ref{ssec:baselines} would predict of two procedures optimizing nearly the same
objective. We also note that the conditions below were identified
\emph{post hoc} from these same experiments, so they are best read as a
mechanistic hypothesis about where density binds---consistent with every
contrast we ran, and with the \aic{} mediation---rather than as a
pre-registered prediction confirmed out of sample. The comparison against
deployed top-$k$ packing is far less delicate: it is positive and significant in
every HotpotQA configuration we test (three reader families, four scales, four
budgets) and on ExpertQA. It too is null where retrieval or task structure
leaves nothing to pack well---MuSiQue $-0.003$, 2Wiki $+0.016$, CovidQA
$+0.010$---which is the same boundary, arrived at from the other direction: when
the evidence is not there, no packing objective can manufacture it.

\subsection{Nothing to assemble}
\label{ssec:cond1}
The objective can only pay when the budget forces a choice among
\emph{complementary} evidence. Two settings remove that precondition, and the
edge vanishes in both. On RAGBench \citep{friel2024ragbench} CovidQA
($n{=}246$) and ExpertQA ($n{=}203$)
the same factorial at budget 160 shows no significant submod--focused gap
(CovidQA $-0.010$ F1, $p{=}0.30$; ExpertQA $+0.005$, $p{=}0.15$): these are
single-pass tasks whose context is largely all-gold, so there is no multi-hop
structure to assemble. On MuSiQue the task \emph{is} multi-hop, but retrieval is
the bottleneck---recall@5${=}0.506$ yet \textbf{all-gold@5${=}0.184$}, only 18\%
of questions with all gold retrieved---and the gap is again null ($+0.011$ F1,
$p{=}0.10$, three seeds).\footnote{Not a matter of depth: tripling retrieval
(top-$k$ $5{\to}12$, nodes $48{\to}64$, expand $5{\to}8$) leaves all-gold@5
unchanged ($0.184$) and the gap null, so the fix is an iterative multi-hop
retriever \citep{trivedi2023interleaving,xiong2021answering}, not a bigger
pool.} The packer cannot assemble evidence that is absent or was never
surfaced---yet the diagnostic still governs quality in both, and is in fact
\emph{strongest} on MuSiQue ($r{=}0.54$, Table~\ref{tab:generalization}).

\paragraph{Budget.} The edge likewise needs a budget that binds without starving
the context. Sweeping $B\in\{96,128,160,224\}$ (three seeds), the
submod$-$focused gap is an inverted-U significant only at ${\approx}160$, while
submod beats naive packing at \emph{every} budget ($+0.036$ to $+0.058$, all
$p\le0.001$) and \texttt{submod@160} matches \texttt{focused@224} at
${\approx}30\%$ fewer tokens (Table~\ref{tab:app_budget}).

\subsection{A reader that is the bottleneck}
\label{ssec:reader}

\begin{figure}[t]
\centering
\includegraphics[width=\columnwidth]{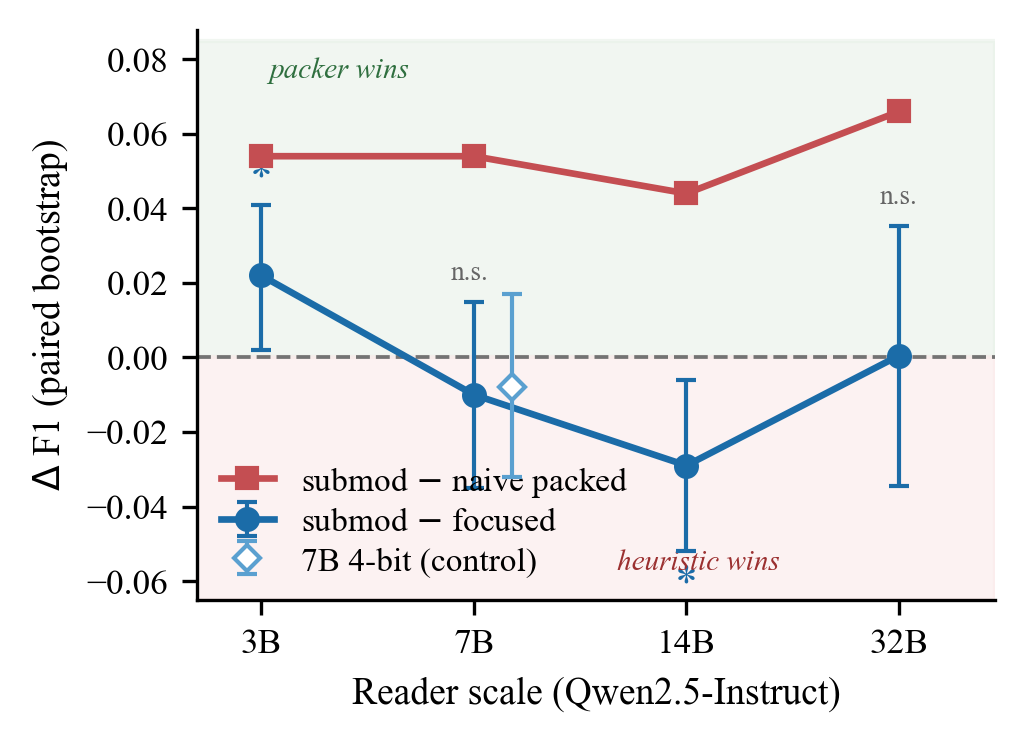}
\caption{Reader-scale ladder (HotpotQA, budget 160; 3B three-seed, 7B/14B
two-seed, 32B single-seed). The packer's edge over the \emph{focused heuristic}
(blue) is positive at 3B, null at 7B, significantly \emph{negative} at 14B
($^{*}p{<}0.05$), and back at parity by 32B; the 7B fp16-vs-4-bit control
(hollow diamond) overlaps the fp16 point, so the trend is scale, not
quantization. The edge over \emph{naive packing} (red) stays significantly
positive at every rung ($+0.044$ to $+0.066$), with no systematic trend in
magnitude across scale.}
\label{fig:ladder}
\end{figure}

The sharpest objection to \S\ref{sec:results} is scaling: \emph{a stronger
reader recovers the answer from messier context, so a packer that merely tidies
it is irrelevant at scale.} We trace the advantage along a \textbf{reader-scale
ladder}---Qwen2.5 at 3B, 7B, 14B, and 32B---re-running the exact factorial and
changing only the reader. Since 14B/32B need 4-bit (NF4) quantization
\citep{dettmers2023qlora} to fit dual T4s, a same-size \textbf{precision control}
(7B fp16 \emph{and} 4-bit) attributes any trend to scale, not quantization.

Figure~\ref{fig:ladder} (contrasts and CIs in App.~\ref{app:reader}) tells a
clean story. At 3B the packer beats the focused heuristic ($+0.022$); at 7B a
symmetric null; at 14B the \textbf{heuristic significantly beats it} ($-0.029$,
$p{=}0.013$); at 32B they are back at parity. The edge over the heuristic is thus
absorbed once the reader passes 3B and never returns---a dip at 14B, parity at
32B---not a runaway cost.\footnote{A 7B fp16-vs-4-bit control rules out
quantization as the driver: the two overlap ($-0.010$ vs.\ $-0.008$, same null
and best policy).} The win over \emph{naive} packing is not absorbed at all:
\submod{} still packs more gold (coverage ${\approx}0.78$ vs.\ $0.73$) and still
beats it at every rung ($+0.044$ to $+0.066$ F1, $p\le0.001$), with no
systematic trend in magnitude---we read this as scale-invariance, not growth,
since the top rung is single-seed. Reader capability is a \emph{mediator}: once a reader
can extract the answer from the focused pack, denser gold stops buying accuracy,
and by 32B the packing overhead stops hurting too.

\paragraph{Synthesis.}
The heuristic-beating edge therefore lives where evidence density---not task
structure, retrieval, or reading capacity---is the binding constraint: multi-hop
questions whose complementary evidence is actually retrieved, a budget that
binds without starving the context, and a reader small enough to need the help.
The two bounds differ in kind: where there is nothing to assemble the mechanism
never engages, whereas a large reader engages it but stops \emph{needing} the
result. The \textbf{diagnostic is the unifying variable}---each bound is a
distinct reason the packer cannot raise \aic{}, and accuracy tracks \aic{}
throughout.

\section{Discussion}
\label{sec:discussion}

\paragraph{What parity with the heuristic does and does not mean.}
Across most settings the packer and the focused heuristic are indistinguishable,
and we think that is the more interesting reading of the two. The heuristic
encodes, by hand, two of the four terms our objective derives
(\S\ref{ssec:baselines}); that a principled formulation matches a carefully
hand-tuned one is evidence the objective is \emph{right}, not that it is
useless---and the formulation supplies what hand-tuning does not: cost
normalization, an approximation guarantee, and no per-task tuning. The practical
statement for a practitioner is correspondingly modest and, we think, more
useful than a headline number: if you are already running a well-tuned
query-focused heuristic, principled packing will roughly match it; if you are
running top-$k$-and-truncate or a prompt compressor---as most deployed systems
are---it is a substantial and consistent gain.

\section{Conclusion}
\label{sec:conclusion}

Budget-constrained multi-hop RAG is bottlenecked not by how many gold documents
are retrieved but by whether the answer survives packing into the reader
context. Our main contribution is a measurement for that: \textbf{\aic{}}, which
predicts answer quality better than retrieval recall across five datasets and
carries information recall does not ($\Delta R^2{=}{+}0.17$ to ${+}0.27$ on three
multi-hop datasets), confirmed by two independent interventions---a packing
change that raises coverage but not \aic{} leaves accuracy flat, and a compressor
that destroys the answer span lowers both. Acting on the diagnostic pays: a
\textbf{budgeted submodular packer} that optimizes it beats deployed
top-$k$ packing and LLMLingua-2 compression at equal-or-lower token cost, across
three reader families and 3B--32B readers. Against a hand-tuned query-focused
heuristic---which we show approximates the same objective---it reaches parity in
most settings and wins where evidence density is the binding constraint. Taken
together: a validated measurement for budgeted RAG, evidence that optimizing it
helps, and an honest account of the one comparison where a well-tuned heuristic
is already close.

\section*{Limitations}
\label{sec:limitations}

The \emph{edge over the focused heuristic} is demonstrated on \emph{one} dataset
(HotpotQA), one budget regime, and one reader family (Qwen2.5-3B). Because we
report 14 such contrasts, of which one is positive and one negative at
$p{<}0.05$, that single positive result is roughly what a family of tests would
produce under no effect; we therefore treat it as a hypothesis about where
density binds---supported by the \aic{} mediation and by the direction of the
remaining nulls, but not established---and would want an out-of-sample,
pre-registered replication before calling it settled. Nothing else in the paper
rests on it: the diagnostic's validity and the comparisons against deployed
top-$k$ packing and compression are each significant across many more settings
than they are null. The reader-scale ladder spans 3B/7B/14B/32B but only within
Qwen2.5 and a single embedder (\texttt{bge-small-en}), and the cross-family
check is at 3B only, so the \emph{interaction} of scale and family, stronger or
instruction-tuned retrievers, and readers beyond 32B are untested. The budget
sweep and MuSiQue are now three-seed; the 7B/14B ladder rungs are two-seed and
the 32B rung is single-seed (compute-bound). The graded \aic{} variant
(\S\ref{ssec:generalization}) handles long free-form answers, but it is still a
lexical overlap measure; a fully semantic/entailment version is future work. The
ACE graph construction is heuristic, so the ``packing substitutes for graph
compression'' reading should be taken with that caveat. Finally, we measure
EM/F1 and context properties, not attribution faithfulness
\citep{es2024ragas,saadfalcon2024ares}; a faithfulness-aware version of \aic{}
is left to future work.

\bibliography{custom}

@inproceedings{lewis2020retrieval,
  title     = {Retrieval-Augmented Generation for Knowledge-Intensive {NLP} Tasks},
  author    = {Lewis, Patrick and Perez, Ethan and Piktus, Aleksandra and Petroni, Fabio and Karpukhin, Vladimir and Goyal, Naman and K{\"u}ttler, Heinrich and Lewis, Mike and Yih, Wen-tau and Rockt{\"a}schel, Tim and Riedel, Sebastian and Kiela, Douwe},
  booktitle = {Advances in Neural Information Processing Systems (NeurIPS)},
  volume    = {33},
  pages     = {9459--9474},
  year      = {2020},
}

@inproceedings{guu2020realm,
  title     = {{REALM}: Retrieval-Augmented Language Model Pre-Training},
  author    = {Guu, Kelvin and Lee, Kenton and Tung, Zora and Pasupat, Panupong and Chang, Ming-Wei},
  booktitle = {Proceedings of the 37th International Conference on Machine Learning (ICML)},
  pages     = {3929--3938},
  year      = {2020},
}

@inproceedings{karpukhin2020dense,
  title     = {Dense Passage Retrieval for Open-Domain Question Answering},
  author    = {Karpukhin, Vladimir and O{\u{g}}uz, Barlas and Min, Sewon and Lewis, Patrick and Wu, Ledell and Edunov, Sergey and Chen, Danqi and Yih, Wen-tau},
  booktitle = {Proceedings of the 2020 Conference on Empirical Methods in Natural Language Processing (EMNLP)},
  pages     = {6769--6781},
  year      = {2020},
}

@inproceedings{izacard2021leveraging,
  title     = {Leveraging Passage Retrieval with Generative Models for Open Domain Question Answering},
  author    = {Izacard, Gautier and Grave, Edouard},
  booktitle = {Proceedings of the 16th Conference of the European Chapter of the Association for Computational Linguistics (EACL)},
  pages     = {874--880},
  year      = {2021},
}

@article{izacard2023atlas,
  title   = {Atlas: Few-shot Learning with Retrieval Augmented Language Models},
  author  = {Izacard, Gautier and Lewis, Patrick and Lomeli, Maria and Hosseini, Lucas and Petroni, Fabio and Schick, Timo and Dwivedi-Yu, Jane and Joulin, Armand and Riedel, Sebastian and Grave, Edouard},
  journal = {Journal of Machine Learning Research (JMLR)},
  volume  = {24},
  number  = {251},
  pages   = {1--43},
  year    = {2023},
}

@inproceedings{borgeaud2022improving,
  title     = {Improving Language Models by Retrieving from Trillions of Tokens},
  author    = {Borgeaud, Sebastian and Mensch, Arthur and Hoffmann, Jordan and Cai, Trevor and Rutherford, Eliza and Millican, Katie and van den Driessche, George and Lespiau, Jean-Baptiste and Damoc, Bogdan and Clark, Aidan and de las Casas, Diego and Guy, Aurelia and Menick, Jacob and Ring, Roman and Hennigan, Tom and Huang, Saffron and Maggiore, Loren and Jones, Chris and Cassirer, Albin and Brock, Andy and Paganini, Michela and Irving, Geoffrey and Vinyals, Oriol and Osindero, Simon and Simonyan, Karen and Rae, Jack and Elsen, Erich and Sifre, Laurent},
  booktitle = {Proceedings of the 39th International Conference on Machine Learning (ICML)},
  pages     = {2206--2240},
  year      = {2022},
}

@article{ram2023incontext,
  title   = {In-Context Retrieval-Augmented Language Models},
  author  = {Ram, Ori and Levine, Yoav and Dalmedigos, Itay and Muhlgay, Dor and Shashua, Amnon and Leyton-Brown, Kevin and Shoham, Yoav},
  journal = {Transactions of the Association for Computational Linguistics (TACL)},
  volume  = {11},
  pages   = {1316--1331},
  year    = {2023},
}

@inproceedings{shi2024replug,
  title     = {{REPLUG}: Retrieval-Augmented Black-Box Language Models},
  author    = {Shi, Weijia and Min, Sewon and Yasunaga, Michihiro and Seo, Minjoon and James, Richard and Lewis, Mike and Zettlemoyer, Luke and Yih, Wen-tau},
  booktitle = {Proceedings of the 2024 Conference of the North American Chapter of the Association for Computational Linguistics (NAACL)},
  pages     = {8371--8384},
  year      = {2024},
}

@inproceedings{asai2024selfrag,
  title     = {Self-{RAG}: Learning to Retrieve, Generate, and Critique through Self-Reflection},
  author    = {Asai, Akari and Wu, Zeqiu and Wang, Yizhong and Sil, Avirup and Hajishirzi, Hannaneh},
  booktitle = {International Conference on Learning Representations (ICLR)},
  year      = {2024},
}

@inproceedings{lin2024radit,
  title     = {{RA-DIT}: Retrieval-Augmented Dual Instruction Tuning},
  author    = {Lin, Xi Victoria and Chen, Xilun and Chen, Mingda and Shi, Weijia and Lomeli, Maria and James, Rich and Rodriguez, Pedro and Kahn, Jacob and Szilvasy, Gergely and Lewis, Mike and Zettlemoyer, Luke and Yih, Scott},
  booktitle = {International Conference on Learning Representations (ICLR)},
  year      = {2024},
}

@article{gao2023retrieval,
  title   = {Retrieval-Augmented Generation for Large Language Models: A Survey},
  author  = {Gao, Yunfan and Xiong, Yun and Gao, Xinyu and Jia, Kangxiang and Pan, Jinliu and Bi, Yuxi and Dai, Yi and Sun, Jiawei and Wang, Meng and Wang, Haofen},
  journal = {arXiv preprint arXiv:2312.10997},
  year    = {2023},
}

@inproceedings{yang2018hotpotqa,
  title     = {{HotpotQA}: A Dataset for Diverse, Explainable Multi-hop Question Answering},
  author    = {Yang, Zhilin and Qi, Peng and Zhang, Saizheng and Bengio, Yoshua and Cohen, William and Salakhutdinov, Ruslan and Manning, Christopher D.},
  booktitle = {Proceedings of the 2018 Conference on Empirical Methods in Natural Language Processing (EMNLP)},
  pages     = {2369--2380},
  year      = {2018},
}

@article{trivedi2022musique,
  title   = {{MuSiQue}: Multihop Questions via Single-hop Question Composition},
  author  = {Trivedi, Harsh and Balasubramanian, Niranjan and Khot, Tushar and Sabharwal, Ashish},
  journal = {Transactions of the Association for Computational Linguistics (TACL)},
  volume  = {10},
  pages   = {539--554},
  year    = {2022},
}

@inproceedings{ho2020constructing,
  title     = {Constructing A Multi-hop {QA} Dataset for Comprehensive Evaluation of Reasoning Steps},
  author    = {Ho, Xanh and Duong Nguyen, Anh-Khoa and Sugawara, Saku and Aizawa, Akiko},
  booktitle = {Proceedings of the 28th International Conference on Computational Linguistics (COLING)},
  pages     = {6609--6625},
  year      = {2020},
}

@article{welbl2018constructing,
  title   = {Constructing Datasets for Multi-hop Reading Comprehension Across Documents},
  author  = {Welbl, Johannes and Stenetorp, Pontus and Riedel, Sebastian},
  journal = {Transactions of the Association for Computational Linguistics (TACL)},
  volume  = {6},
  pages   = {287--302},
  year    = {2018},
}

@inproceedings{trivedi2023interleaving,
  title     = {Interleaving Retrieval with Chain-of-Thought Reasoning for Knowledge-Intensive Multi-Step Questions},
  author    = {Trivedi, Harsh and Balasubramanian, Niranjan and Khot, Tushar and Sabharwal, Ashish},
  booktitle = {Proceedings of the 61st Annual Meeting of the Association for Computational Linguistics (ACL)},
  pages     = {10014--10037},
  year      = {2023},
}

@inproceedings{press2023measuring,
  title     = {Measuring and Narrowing the Compositionality Gap in Language Models},
  author    = {Press, Ofir and Zhang, Muru and Min, Sewon and Schmidt, Ludwig and Smith, Noah A. and Lewis, Mike},
  booktitle = {Findings of the Association for Computational Linguistics: EMNLP 2023},
  pages     = {5687--5711},
  year      = {2023},
}

@inproceedings{xiong2021answering,
  title     = {Answering Complex Open-Domain Questions with Multi-Hop Dense Retrieval},
  author    = {Xiong, Wenhan and Li, Xiang Lorraine and Iyer, Srinivasan and Du, Jingfei and Lewis, Patrick and Wang, William Yang and Mehdad, Yashar and Yih, Wen-tau and Riedel, Sebastian and Kiela, Douwe and O{\u{g}}uz, Barlas},
  booktitle = {International Conference on Learning Representations (ICLR)},
  year      = {2021},
}

@article{khattab2022demonstrate,
  title   = {Demonstrate-Search-Predict: Composing Retrieval and Language Models for Knowledge-Intensive {NLP}},
  author  = {Khattab, Omar and Santhanam, Keshav and Li, Xiang Lisa and Hall, David and Liang, Percy and Potts, Christopher and Zaharia, Matei},
  journal = {arXiv preprint arXiv:2212.14024},
  year    = {2022},
}

@inproceedings{jiang2023active,
  title     = {Active Retrieval Augmented Generation},
  author    = {Jiang, Zhengbao and Xu, Frank and Gao, Luyu and Sun, Zhiqing and Liu, Qian and Dwivedi-Yu, Jane and Yang, Yiming and Callan, Jamie and Neubig, Graham},
  booktitle = {Proceedings of the 2023 Conference on Empirical Methods in Natural Language Processing (EMNLP)},
  pages     = {7969--7992},
  year      = {2023},
}

@article{robertson2009probabilistic,
  title   = {The Probabilistic Relevance Framework: {BM25} and Beyond},
  author  = {Robertson, Stephen and Zaragoza, Hugo},
  journal = {Foundations and Trends in Information Retrieval},
  volume  = {3},
  number  = {4},
  pages   = {333--389},
  year    = {2009},
}

@inproceedings{reimers2019sentence,
  title     = {Sentence-{BERT}: Sentence Embeddings using Siamese {BERT}-Networks},
  author    = {Reimers, Nils and Gurevych, Iryna},
  booktitle = {Proceedings of the 2019 Conference on Empirical Methods in Natural Language Processing (EMNLP-IJCNLP)},
  pages     = {3982--3992},
  year      = {2019},
}

@article{nogueira2019passage,
  title   = {Passage Re-ranking with {BERT}},
  author  = {Nogueira, Rodrigo and Cho, Kyunghyun},
  journal = {arXiv preprint arXiv:1901.04085},
  year    = {2019},
}

@inproceedings{khattab2020colbert,
  title     = {{ColBERT}: Efficient and Effective Passage Search via Contextualized Late Interaction over {BERT}},
  author    = {Khattab, Omar and Zaharia, Matei},
  booktitle = {Proceedings of the 43rd International ACM SIGIR Conference on Research and Development in Information Retrieval (SIGIR)},
  pages     = {39--48},
  year      = {2020},
}

@inproceedings{xiao2024cpack,
  title     = {{C-Pack}: Packed Resources For General Chinese Embeddings},
  author    = {Xiao, Shitao and Liu, Zheng and Zhang, Peitian and Muennighoff, Niklas and Lian, Defu and Nie, Jian-Yun},
  booktitle = {Proceedings of the 47th International ACM SIGIR Conference on Research and Development in Information Retrieval (SIGIR)},
  pages     = {641--649},
  year      = {2024},
}

@inproceedings{muennighoff2023mteb,
  title     = {{MTEB}: Massive Text Embedding Benchmark},
  author    = {Muennighoff, Niklas and Tazi, Nouamane and Magne, Lo{\"i}c and Reimers, Nils},
  booktitle = {Proceedings of the 17th Conference of the European Chapter of the Association for Computational Linguistics (EACL)},
  pages     = {2014--2037},
  year      = {2023},
}

@inproceedings{thakur2021beir,
  title     = {{BEIR}: A Heterogeneous Benchmark for Zero-shot Evaluation of Information Retrieval Models},
  author    = {Thakur, Nandan and Reimers, Nils and R{\"u}ckl{\'e}, Andreas and Srivastava, Abhishek and Gurevych, Iryna},
  booktitle = {Advances in Neural Information Processing Systems (NeurIPS), Datasets and Benchmarks Track},
  year      = {2021},
}

@inproceedings{carbonell1998use,
  title     = {The Use of {MMR}, Diversity-Based Reranking for Reordering Documents and Producing Summaries},
  author    = {Carbonell, Jaime and Goldstein, Jade},
  booktitle = {Proceedings of the 21st Annual International ACM SIGIR Conference on Research and Development in Information Retrieval (SIGIR)},
  pages     = {335--336},
  year      = {1998},
}

@article{liu2024lost,
  title   = {Lost in the Middle: How Language Models Use Long Contexts},
  author  = {Liu, Nelson F. and Lin, Kevin and Hewitt, John and Paranjape, Ashwin and Bevilacqua, Michele and Petroni, Fabio and Liang, Percy},
  journal = {Transactions of the Association for Computational Linguistics (TACL)},
  volume  = {12},
  pages   = {157--173},
  year    = {2024},
}

@inproceedings{xu2024recomp,
  title     = {{RECOMP}: Improving Retrieval-Augmented {LM}s with Context Compression and Selective Augmentation},
  author    = {Xu, Fangyuan and Shi, Weijia and Choi, Eunsol},
  booktitle = {International Conference on Learning Representations (ICLR)},
  year      = {2024},
}

@inproceedings{jiang2023llmlingua,
  title     = {{LLMLingua}: Compressing Prompts for Accelerated Inference of Large Language Models},
  author    = {Jiang, Huiqiang and Wu, Qianhui and Lin, Chin-Yew and Yang, Yuqing and Qiu, Lili},
  booktitle = {Proceedings of the 2023 Conference on Empirical Methods in Natural Language Processing (EMNLP)},
  pages     = {13358--13376},
  year      = {2023},
}

@inproceedings{pan2024llmlingua2,
  title     = {{LLMLingua-2}: Data Distillation for Efficient and Faithful Task-Agnostic Prompt Compression},
  author    = {Pan, Zhuoshi and Wu, Qianhui and Jiang, Huiqiang and Xia, Menglin and Luo, Xufang and Zhang, Jue and Lin, Qingwei and R{\"u}hle, Victor and Yang, Yuqing and Lin, Chin-Yew and Zhao, H. Vicky and Qiu, Lili and Zhang, Dongmei},
  booktitle = {Findings of the Association for Computational Linguistics (ACL Findings)},
  pages     = {963--981},
  year      = {2024},
}

@inproceedings{li2023compressing,
  title     = {Compressing Context to Enhance Inference Efficiency of Large Language Models},
  author    = {Li, Yucheng and Dong, Bo and Guerin, Frank and Lin, Chenghua},
  booktitle = {Proceedings of the 2023 Conference on Empirical Methods in Natural Language Processing (EMNLP)},
  pages     = {6342--6353},
  year      = {2023},
}

@article{wang2023learning,
  title   = {Learning to Filter Context for Retrieval-Augmented Generation},
  author  = {Wang, Zhiruo and Araki, Jun and Jiang, Zhengbao and Parvez, Md Rizwan and Neubig, Graham},
  journal = {arXiv preprint arXiv:2311.08377},
  year    = {2023},
}

@inproceedings{yoran2024making,
  title     = {Making Retrieval-Augmented Language Models Robust to Irrelevant Context},
  author    = {Yoran, Ori and Wolfson, Tomer and Ram, Ori and Berant, Jonathan},
  booktitle = {International Conference on Learning Representations (ICLR)},
  year      = {2024},
}

@inproceedings{bai2024longbench,
  title     = {{LongBench}: A Bilingual, Multitask Benchmark for Long Context Understanding},
  author    = {Bai, Yushi and Lv, Xin and Zhang, Jiajie and Lyu, Hongchang and Tang, Jiankai and Huang, Zhidian and Du, Zhengxiao and Liu, Xiao and Zeng, Aohan and Hou, Lei and Dong, Yuxiao and Tang, Jie and Li, Juanzi},
  booktitle = {Proceedings of the 62nd Annual Meeting of the Association for Computational Linguistics (ACL)},
  pages     = {3119--3137},
  year      = {2024},
}

@inproceedings{xu2024retrieval,
  title     = {Retrieval Meets Long Context Large Language Models},
  author    = {Xu, Peng and Ping, Wei and Wu, Xianchao and McAfee, Lawrence and Zhu, Chen and Liu, Zihan and Subramanian, Sandeep and Bakhturina, Evelina and Shoeybi, Mohammad and Catanzaro, Bryan},
  booktitle = {International Conference on Learning Representations (ICLR)},
  year      = {2024},
}

@inproceedings{lin2011class,
  title     = {A Class of Submodular Functions for Document Summarization},
  author    = {Lin, Hui and Bilmes, Jeff},
  booktitle = {Proceedings of the 49th Annual Meeting of the Association for Computational Linguistics: Human Language Technologies (ACL-HLT)},
  pages     = {510--520},
  year      = {2011},
}

@inproceedings{lin2010multidocument,
  title     = {Multi-document Summarization via Budgeted Maximization of Submodular Functions},
  author    = {Lin, Hui and Bilmes, Jeff},
  booktitle = {Human Language Technologies: The 2010 Annual Conference of the North American Chapter of the Association for Computational Linguistics (NAACL-HLT)},
  pages     = {912--920},
  year      = {2010},
}

@article{nemhauser1978analysis,
  title   = {An Analysis of Approximations for Maximizing Submodular Set Functions---{I}},
  author  = {Nemhauser, George L. and Wolsey, Laurence A. and Fisher, Marshall L.},
  journal = {Mathematical Programming},
  volume  = {14},
  number  = {1},
  pages   = {265--294},
  year    = {1978},
}

@incollection{krause2014submodular,
  title     = {Submodular Function Maximization},
  author    = {Krause, Andreas and Golovin, Daniel},
  booktitle = {Tractability: Practical Approaches to Hard Problems},
  pages     = {71--104},
  publisher = {Cambridge University Press},
  year      = {2014},
}

@article{bilmes2022submodularity,
  title   = {Submodularity In Machine Learning and Artificial Intelligence},
  author  = {Bilmes, Jeff},
  journal = {arXiv preprint arXiv:2202.00132},
  year    = {2022},
}

@article{qwen2025qwen25,
  title   = {{Qwen2.5} Technical Report},
  author  = {{Qwen Team}},
  journal = {arXiv preprint arXiv:2412.15115},
  year    = {2024},
}

@article{touvron2023llama,
  title   = {{Llama 2}: Open Foundation and Fine-Tuned Chat Models},
  author  = {Touvron, Hugo and Martin, Louis and Stone, Kevin and Albert, Peter and Almahairi, Amjad and Babaei, Yasmine and Bashlykov, Nikolay and Batra, Soumya and Bhargava, Prajjwal and Bhosale, Shruti and Bikel, Dan and Blecher, Lukas and Canton Ferrer, Cristian and Chen, Moya and Cucurull, Guillem and Esiobu, David and Fernandes, Jude and Fu, Jeremy and Fu, Wenyin and Fuller, Brian and Gao, Cynthia and Goswami, Vedanuj and Goyal, Naman and Hartshorn, Anthony and Hosseini, Saghar and Hou, Rui and Inan, Hakan and Kardas, Marcin and Kerkez, Viktor and Khabsa, Madian and Kloumann, Isabel and Korenev, Artem and Koura, Punit Singh and Lachaux, Marie-Anne and Lavril, Thibaut and Lee, Jenya and Liskovich, Diana and Lu, Yinghai and Mao, Yuning and Martinet, Xavier and Mihaylov, Todor and Mishra, Pushkar and Molybog, Igor and Nie, Yixin and Poulton, Andrew and Reizenstein, Jeremy and Rungta, Rashi and Saladi, Kalyan and Schelten, Alan and Silva, Ruan and Smith, Eric Michael and Subramanian, Ranjan and Tan, Xiaoqing Ellen and Tang, Binh and Taylor, Ross and Williams, Adina and Kuan, Jian Xiang and Xu, Puxin and Yan, Zheng and Zarov, Iliyan and Zhang, Yuchen and Fan, Angela and Kambadur, Melanie and Narang, Sharan and Rodriguez, Aurelien and Stojnic, Robert and Edunov, Sergey and Scialom, Thomas},
  journal = {arXiv preprint arXiv:2307.09288},
  year    = {2023},
}

@inproceedings{brown2020language,
  title     = {Language Models are Few-Shot Learners},
  author    = {Brown, Tom B. and Mann, Benjamin and Ryder, Nick and Subbiah, Melanie and Kaplan, Jared and Dhariwal, Prafulla and Neelakantan, Arvind and Shyam, Pranav and Sastry, Girish and Askell, Amanda and Agarwal, Sandhini and Herbert-Voss, Ariel and Krueger, Gretchen and Henighan, Tom and Child, Rewon and Ramesh, Aditya and Ziegler, Daniel M. and Wu, Jeffrey and Winter, Clemens and Hesse, Christopher and Chen, Mark and Sigler, Eric and Litwin, Mateusz and Gray, Scott and Chess, Benjamin and Clark, Jack and Berner, Christopher and McCandlish, Sam and Radford, Alec and Sutskever, Ilya and Amodei, Dario},
  booktitle = {Advances in Neural Information Processing Systems (NeurIPS)},
  volume    = {33},
  pages     = {1877--1901},
  year      = {2020},
}

@inproceedings{dettmers2023qlora,
  title     = {{QLoRA}: Efficient Finetuning of Quantized {LLM}s},
  author    = {Dettmers, Tim and Pagnoni, Artidoro and Holtzman, Ari and Zettlemoyer, Luke},
  booktitle = {Advances in Neural Information Processing Systems (NeurIPS)},
  year      = {2023},
}

@inproceedings{dettmers2022llmint8,
  title     = {{LLM.int8()}: 8-bit Matrix Multiplication for Transformers at Scale},
  author    = {Dettmers, Tim and Lewis, Mike and Belkada, Younes and Zettlemoyer, Luke},
  booktitle = {Advances in Neural Information Processing Systems (NeurIPS)},
  year      = {2022},
}

@inproceedings{rajpurkar2016squad,
  title     = {{SQuAD}: 100,000+ Questions for Machine Comprehension of Text},
  author    = {Rajpurkar, Pranav and Zhang, Jian and Lopyrev, Konstantin and Liang, Percy},
  booktitle = {Proceedings of the 2016 Conference on Empirical Methods in Natural Language Processing (EMNLP)},
  pages     = {2383--2392},
  year      = {2016},
}

@inproceedings{es2024ragas,
  title     = {{RAGAS}: Automated Evaluation of Retrieval Augmented Generation},
  author    = {Es, Shahul and James, Jithin and Espinosa Anke, Luis and Schockaert, Steven},
  booktitle = {Proceedings of the 18th Conference of the European Chapter of the Association for Computational Linguistics (EACL): System Demonstrations},
  pages     = {150--158},
  year      = {2024},
}

@inproceedings{saadfalcon2024ares,
  title     = {{ARES}: An Automated Evaluation Framework for Retrieval-Augmented Generation Systems},
  author    = {Saad-Falcon, Jon and Khattab, Omar and Potts, Christopher and Zaharia, Matei},
  booktitle = {Proceedings of the 2024 Conference of the North American Chapter of the Association for Computational Linguistics (NAACL)},
  pages     = {338--354},
  year      = {2024},
}

@inproceedings{chen2024benchmarking,
  title     = {Benchmarking Large Language Models in Retrieval-Augmented Generation},
  author    = {Chen, Jiawei and Lin, Hongyu and Han, Xianpei and Sun, Le},
  booktitle = {Proceedings of the AAAI Conference on Artificial Intelligence},
  volume    = {38},
  pages     = {17754--17762},
  year      = {2024},
}

@inproceedings{petroni2021kilt,
  title     = {{KILT}: a Benchmark for Knowledge Intensive Language Tasks},
  author    = {Petroni, Fabio and Piktus, Aleksandra and Fan, Angela and Lewis, Patrick and Yazdani, Majid and De Cao, Nicola and Thorne, James and Jernite, Yacine and Karpukhin, Vladimir and Maillard, Jean and Plachouras, Vassilis and Rockt{\"a}schel, Tim and Riedel, Sebastian},
  booktitle = {Proceedings of the 2021 Conference of the North American Chapter of the Association for Computational Linguistics (NAACL)},
  pages     = {2523--2544},
  year      = {2021},
}

@inproceedings{mallen2023trust,
  title     = {When Not to Trust Language Models: Investigating the Effectiveness of Parametric and Non-Parametric Memories},
  author    = {Mallen, Alex and Asai, Akari and Zhong, Victor and Das, Rajarshi and Khashabi, Daniel and Hajishirzi, Hannaneh},
  booktitle = {Proceedings of the 61st Annual Meeting of the Association for Computational Linguistics (ACL)},
  pages     = {9802--9822},
  year      = {2023},
}

@inproceedings{wei2022chain,
  title     = {Chain-of-Thought Prompting Elicits Reasoning in Large Language Models},
  author    = {Wei, Jason and Wang, Xuezhi and Schuurmans, Dale and Bosma, Maarten and Ichter, Brian and Xia, Fei and Chi, Ed H. and Le, Quoc V. and Zhou, Denny},
  booktitle = {Advances in Neural Information Processing Systems (NeurIPS)},
  year      = {2022},
}

@inproceedings{shao2023enhancing,
  title     = {Enhancing Retrieval-Augmented Large Language Models with Iterative Retrieval-Generation Synergy},
  author    = {Shao, Zhihong and Gong, Yeyun and Shen, Yelong and Huang, Minlie and Duan, Nan and Chen, Weizhu},
  booktitle = {Findings of the Association for Computational Linguistics: EMNLP 2023},
  pages     = {9248--9274},
  year      = {2023},
}

@article{friel2024ragbench,
  title   = {{RAGBench}: Explainable Benchmark for Retrieval-Augmented Generation Systems},
  author  = {Friel, Robert and Belyi, Masha and Sanyal, Atindriyo},
  journal = {arXiv preprint arXiv:2407.11005},
  year    = {2024},
}

@misc{falcon3,
  title        = {The {Falcon} 3 Family of Open Models},
  author       = {{Falcon-LLM Team}},
  year         = {2024},
  howpublished = {\url{https://huggingface.co/tiiuae/Falcon3-3B-Instruct}},
  note         = {Technology Innovation Institute},
}

@article{abdin2024phi3,
  title   = {Phi-3 Technical Report: A Highly Capable Language Model Locally on Your Phone},
  author  = {Abdin, Marah and Aneja, Jyoti and Awadalla, Hany and Awadallah, Ahmed and Awan, Ammar Ahmad and Bach, Nguyen and Bahree, Amit and Bakhtiari, Arash and Bao, Jianmin and Behl, Harkirat and Benhaim, Alon and Bilenko, Misha and Bjorck, Johan and Bubeck, S{\'e}bastien and Cai, Martin and Cai, Qin and Chaudhary, Vishrav and Chen, Dong and Chen, Dongdong and Chen, Weizhu and Chen, Yen-Chun and Chen, Yi-Ling and Cheng, Hao and Chopra, Parul and Dai, Xiyang and Dixon, Matthew and Eldan, Ronen and Fragoso, Victor and Gao, Jianfeng and Gao, Mei and Gao, Min and Garg, Amit and Del Giorno, Allie and Goswami, Abhishek and Gunasekar, Suriya and Haider, Emman and Hao, Junheng and Hewett, Russell J. and Hu, Wenxiang and Huynh, Jamie and Iter, Dan and Jacobs, Sam Ade and Javaheripi, Mojan and Jin, Xin and Karampatziakis, Nikos and Kauffmann, Piero and Khademi, Mahoud and Kim, Dongwoo and Kim, Young Jin and Kurilenko, Lev and Lee, James R. and Lee, Yin Tat and Li, Yuanzhi and Li, Yunsheng and Liang, Chen and Liden, Lars and Lin, Xihui and Lin, Zeqi and Liu, Ce and Liu, Liyuan and Liu, Mengchen and Liu, Weishung and Liu, Xiaodong and Luo, Chong and Madan, Piyush and Mahmoudzadeh, Ali and Majercak, David and Mazzola, Matt and Mendes, Caio C{\'e}sar Teodoro and Mitra, Arindam and Modi, Hardik and Nguyen, Anh and Norick, Brandon and Patra, Barun and Perez-Becker, Daniel and Portet, Thomas and Pryzant, Reid and Qin, Heyang and Radmilac, Marko and Ren, Liliang and de Rosa, Gustavo and Rosset, Corby and Roy, Sambudha and Ruwase, Olatunji and Saarikivi, Olli and Saied, Amin and Salim, Adil and Santacroce, Michael and Shah, Shital and Shang, Ning and Sharma, Hiteshi and Shen, Yelong and Shukla, Swadheen and Song, Xia and Tanaka, Masahiro and Tupini, Andrea and Vaddamanu, Praneetha and Wang, Chunyu and Wang, Guanhua and Wang, Lijuan and Wang, Shuohang and Wang, Xin and Wang, Yu and Ward, Rachel and Wen, Wen and Witte, Philipp and Wu, Haiping and Wu, Xiaoxia and Wyatt, Michael and Xiao, Bin and Xu, Can and Xu, Jiahang and Xu, Weijian and Xue, Jilong and Yadav, Sonali and Yang, Fan and Yang, Jianwei and Yang, Yifan and Yang, Ziyi and Yu, Donghan and Yuan, Lu and Zhang, Chenruidong and Zhang, Cyril and Zhang, Jianwen and Zhang, Li Lyna and Zhang, Yi and Zhang, Yue and Zhang, Yunan and Zhou, Xiren},
  journal = {arXiv preprint arXiv:2404.14219},
  year    = {2024},
}

@article{sviridenko2004note,
  title   = {A note on maximizing a submodular set function subject to a knapsack constraint},
  author  = {Sviridenko, Maxim},
  journal = {Operations Research Letters},
  volume  = {32},
  number  = {1},
  pages   = {41--43},
  year    = {2004},
}

\appendix

\section{Single-Seed Reference Table}
\label{app:seed42}
Table~\ref{tab:app_seed42} gives the per-policy means underlying the \aic{}
mediation (\S\ref{ssec:correlations}) and decomposition (\S\ref{sec:results}),
computed on seed 42's 500 questions.

\begin{table}[t]
\centering\scriptsize
\setlength{\tabcolsep}{3.2pt}
\begin{tabular}{lcccccc}
\toprule
Policy & F1 & EM & AiC & g-cov & all-g & tok \\
\midrule
\packed{}   & 0.401 & 0.316 & 0.590 & 0.685 & 0.448 & 151.3 \\
\focused{}  & 0.412 & 0.322 & 0.632 & 0.733 & 0.512 & 152.3 \\
\mmrpol{}   & 0.390 & 0.298 & 0.592 & 0.680 & 0.418 & 151.9 \\
\textbf{\submod{}} & \textbf{0.448} & \textbf{0.370} & 0.640 & \textbf{0.779} & \textbf{0.578} & \textbf{145.4} \\
\texttt{ace\_focused} & 0.421 & 0.320 & 0.630 & 0.768 & 0.562 & 150.0 \\
\texttt{ace\_submod}  & 0.401 & 0.308 & 0.612 & 0.758 & 0.550 & 147.4 \\
oracle      & 0.598 & 0.484 & 0.692 & 0.779 & 0.580 & 142.0 \\
\bottomrule
\end{tabular}
\caption{Seed-42 per-policy means, HotpotQA-500, budget 160, 3B reader. AiC =
\aic{}; g-cov${=}$gold-doc reader coverage; all-g${=}$all-gold-in-reader.}
\label{tab:app_seed42}
\end{table}

\begin{table}[t]
\centering\small
\setlength{\tabcolsep}{5pt}
\begin{tabular}{rcccc}
\toprule
Budget & submod F1 & focused F1 & $\Delta$ F1 & $p$ \\
\midrule
 96 & 0.374 & 0.392 & $-0.018$ & 0.08 \\
128 & 0.426 & 0.427 & $-0.001$ & 0.90 \\
\textbf{160} & \textbf{0.451} & \textbf{0.429} & $\mathbf{+0.022}$ & $\mathbf{<.05}$ \\
224 & 0.472 & 0.459 & $+0.013$ & 0.14 \\
\bottomrule
\end{tabular}
\caption{Per-budget F1 underlying the budget sweep (\S\ref{ssec:cond1}), pooled
three-seed $\{42,13,7\}$. The submod$-$focused gap is an inverted-U peaking at
${\approx}160$ (the only significant budget); at the tightest budget it is
mildly negative---too little fits to assemble complementary evidence.}
\label{tab:app_budget}
\end{table}

\section{Reader-Scale Reference Tables}
\label{app:reader}
The packing/diagnostic columns are reader-independent by construction, so they
are identical across rungs; only EM/F1 move. Table~\ref{tab:app_7b} (7B fp16) and
Table~\ref{tab:app_14b} (14B 4-bit) give the two middle rungs in full; the 32B
endpoint (\submod{} $0.438$, \focused{} $0.437$, \packed{} $0.372$; the parity
in Table~\ref{tab:ladder}) shares the same reader-independent packing columns.
The 7B 4-bit control reproduces 7B fp16 (submod$-$focused $-0.007$ F1,
$p{=}0.55$; same best policy on both seeds), with absolute F1 ${\approx}1$--2
points lower (the quantization tax) but the contrast unchanged.
Table~\ref{tab:ladder} gives the full submod$-$focused contrast at every rung
(the source for Figure~\ref{fig:ladder}).

\begin{table}[t]
\centering\small
\setlength{\tabcolsep}{2pt}
\begin{tabular}{lcc}
\toprule
Reader (precision) & submod$-$focused $\Delta$F1 & $p$ \\
\midrule
3B fp16 (\S\ref{sec:results})   & $+0.022\;[+0.002,+0.041]$ & $<$0.05 \\
7B fp16                          & $-0.010\;[-0.035,+0.015]$ & 0.45 \\
7B 4-bit (control)               & $-0.008\;[-0.032,+0.017]$ & 0.55 \\
\textbf{14B 4-bit}               & $\mathbf{-0.029\;[-0.052,-0.006]}$ & \textbf{0.013} \\
32B 4-bit                        & $+0.001\;[-0.035,+0.035]$ & 0.996 \\
\bottomrule
\end{tabular}
\caption{Reader-scale ladder, paired bootstrap (3B three-seed; 7B/14B 2-seed,
$n{=}1{,}000$; 32B single-seed, $n{=}500$). The edge over the heuristic is
absorbed beyond 3B and returns to parity by 32B; the precision control rules out
quantization.}
\label{tab:ladder}
\end{table}

\begin{table}[t]
\centering\small
\setlength{\tabcolsep}{4pt}
\begin{tabular}{lccccc}
\toprule
Policy & F1 & EM & AiC & g-cov & all-g \\
\midrule
\packed{}   & 0.332 & 0.259 & 0.583 & 0.685 & 0.447 \\
\focused{}  & 0.396 & 0.311 & 0.638 & 0.740 & 0.526 \\
\mmrpol{}   & 0.363 & 0.286 & 0.597 & 0.680 & 0.420 \\
\submod{}   & 0.386 & 0.303 & 0.634 & \textbf{0.778} & \textbf{0.582} \\
\texttt{ace\_focused} & 0.390 & 0.303 & 0.624 & 0.760 & 0.553 \\
\texttt{ace\_submod}  & 0.371 & 0.277 & 0.610 & 0.750 & 0.541 \\
oracle      & 0.574 & 0.461 & 0.704 & 0.780 & 0.588 \\
\bottomrule
\end{tabular}
\caption{7B fp16. Per-seed best: seed 42${\to}$\submod{} (0.396); seed
13${\to}$\focused{} (0.407).}
\label{tab:app_7b}
\end{table}

\begin{table}[t]
\centering\small
\setlength{\tabcolsep}{4pt}
\begin{tabular}{lccccc}
\toprule
Policy & F1 & EM & AiC & g-cov & all-g \\
\midrule
\packed{}   & 0.386 & 0.295 & 0.583 & 0.685 & 0.447 \\
\textbf{\focused{}}  & \textbf{0.460} & \textbf{0.356} & 0.638 & 0.740 & 0.526 \\
\mmrpol{}   & 0.413 & 0.311 & 0.597 & 0.680 & 0.420 \\
\submod{}   & 0.431 & 0.329 & 0.634 & \textbf{0.778} & \textbf{0.582} \\
\texttt{ace\_focused} & 0.457 & 0.353 & 0.624 & 0.760 & 0.553 \\
\texttt{ace\_submod}  & 0.416 & 0.323 & 0.610 & 0.750 & 0.541 \\
oracle      & 0.599 & 0.480 & 0.702 & 0.784 & 0.596 \\
\bottomrule
\end{tabular}
\caption{14B 4-bit. Per-seed best: seed 42${\to}$\focused{} (0.459); seed
13${\to}$\texttt{ace\_focused} (0.448). The focused policies are best---the
opposite of 3B---with identical packing underneath.}
\label{tab:app_14b}
\end{table}

\section{2WikiMultiHopQA Interventional Check}
\label{app:2wiki}
3B reader, budget 160, seeds $\{42,13,7\}$, 500 questions. Retrieval gate:
recall@5${=}0.718$, all-gold@5${=}0.43$. Key contrast, pooled 3-seed bootstrap
($n{=}1{,}500$): \submod{}$-$\focused{} ${=}-0.008$ F1 $[-0.027,+0.012]$,
$p{=}0.44$, with coverage $+0.054$ but \aic{} $-0.007$---coverage and \aic{} move
in opposite directions, and F1 follows \aic{}. Conditional F1 is $0.56$ when the
answer is in context versus $0.08$ when not.

\end{document}